\documentclass[10pt,twocolumn,letterpaper]{article}

\usepackage{iccv}
\usepackage{times}
\usepackage{epsfig}
\usepackage{graphicx}
\usepackage{amsmath}
\usepackage{amssymb}

\usepackage{newfloat}
\usepackage{listings}
\usepackage{algorithm}
\usepackage{algorithmic}
\usepackage{multirow}
\usepackage{caption}
\usepackage{subcaption}
\usepackage{pifont}       
\usepackage{bbding}       
\usepackage{fontawesome}
\usepackage{booktabs}
\usepackage{color, colortbl}
\usepackage{xcolor}
\newcommand{\xmark}{\ding{55}}

\usepackage[breaklinks=true,bookmarks=false]{hyperref}

\iccvfinalcopy 

\def\iccvPaperID{****} 

\ificcvfinal\fi

\begin{document}

\title{GaitFormer: Revisiting Intrinsic Periodicity for Gait Recognition}

\author{Qian Wu\\
MEGVII Technology\\
{\tt\small wuqian@megvii.com}
\and
 Ruixuan Xiao\thanks{Work done during the internship at MEGVII Technology.}\\
Zhejiang University\\
{\tt\small xiaoruixuan@zju.edu.cn}
\and
 Kaixin Xu\textsuperscript{$*$}\\
Fudan University\\
{\tt\small  kxxu22@m.fudan.edu.cn}
\and
 Jingcheng Ni\textsuperscript{$*$}\\
Behang University\\
{\tt\small kiranjc@buaa.edu.cn}
\and
Boxun Li\\
MEGVII Technology\\
{\tt\small liboxun@megvii.com}
\and
Ziyao Xu\\
MEGVII Technology\\
{\tt\small xuziyao@megvii.com}
}

\maketitle
\ificcvfinal\thispagestyle{empty}\fi

\begin{abstract}

Gait recognition aims to distinguish different walking patterns by analyzing video-level human silhouettes, rather than relying on appearance information. Previous research on gait recognition has primarily focused on extracting local or global spatial-temporal representations, while overlooking the intrinsic periodic features of gait sequences, which, when fully utilized, can significantly enhance performance. In this work, we propose a plug-and-play strategy, called Temporal Periodic Alignment (TPA), which leverages the periodic nature and fine-grained temporal dependencies of gait patterns. The TPA strategy comprises two key components. The first component is Adaptive Fourier-transform Position Encoding (AFPE), which adaptively converts features and discrete-time signals into embeddings that are sensitive to periodic walking patterns. The second component is the Temporal Aggregation Module (TAM), which separates embeddings into trend and seasonal components, and extracts meaningful temporal correlations to identify primary components, while filtering out random noise. We present a simple and effective baseline method for gait recognition, based on the TPA strategy. Extensive experiments conducted on three popular public datasets (CASIA-B, OU-MVLP, and GREW) demonstrate that our proposed method achieves state-of-the-art performance on multiple benchmark tests.
\end{abstract}

\section{Introduction}
\label{sec:intro}

Biometrics technology aims to identify different people with various physiological characteristics, such as faces, fingerprints, iris, and DNA. 
Compared with other biometrics, human gait, the walking pattern of an individual, could be easily obtained at a distance without the cooperation of interest-subjects. Moreover, it is difficult to camouflage, making it promising for forensic identification, crime prevention, and suspect tracing. 

\begin{figure}[t]
     \includegraphics[width=1\linewidth]{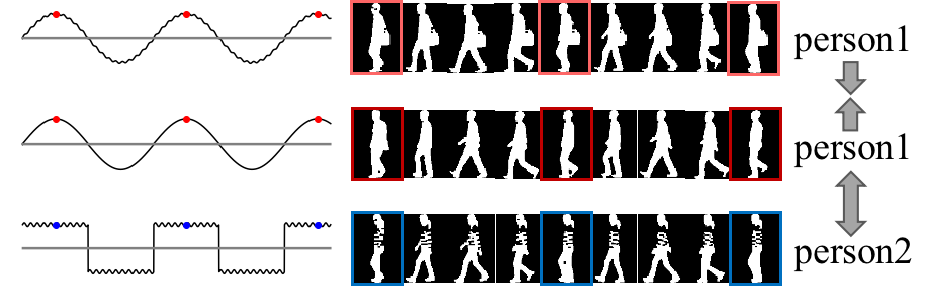}
     \caption{ Based on our observations and statistical analysis, gait sequences exhibit distinct periods that are consistent across most individuals in the CASIA-B dataset. The accompanying waveform illustrates that each gait sequence is composed of several similar periods containing approximately 30 frames. The silhouettes sampled from CASIA-B, arranged from top to bottom, provide a visual representation of these periods.     }
     \label{fig:periodicity}
\end{figure}

However, human gaits are usually captured in complex conditions, including cross-view, pose, carrying, clothing, aging, illumination and occlusions, which have posed a ubiquitous obstacle \cite{connor2018biometric, liao2017pose, yu2006framework, shen2022comprehensive, sepas2022deep}. Thus the core challenge of gait recognition lies in learning discriminative and invariant spatial-temporal features under such unconstrained conditions. From the perspective of temporal information learning, existing approaches can be roughly divided into three types: template-based, image-based, and sequence-based approaches. Template-based approaches directly employ the DNN on single gait templates to extract discriminative representations \cite{DBLP:conf/icb/ShiragaMMEY16,DBLP:conf/icassp/ZhangLMF16}. Image-based methods process 
frames independently with each other, then fuse temporal features with simple operations \cite{DBLP:conf/aaai/ChaoHZF19, DBLP:conf/cvpr/FanPC0HCHLH20, DBLP:conf/iccv/LinZ021}. 
The aforementioned methods rarely take into account the global and local interaction of temporal features.
On the contrary, sequence-based approaches regard sequences as videos and apply LSTM \cite{hochreiter1997long}, 3D-CNN or Set Transformer \cite{litransgait} to fuse every frame for comprehensive motion investigation \cite{DBLP:conf/icpr/FengLL16}. Despite the success, the sequence-based approaches simply apply temporal modules to process the sequence-level information, but rarely employ adaptation to cope with the intrinsic periodicity of gait sequences. At that point, the natural periodic inductive bias gets overlooked, which retains great potential for boosted performance.

As a periodic motion, each gait sequence can be regarded as the combination of several analogous walking periods, as shown in Figure~\ref{fig:periodicity}. Based on the observations of different gait periods, fine-grained similarities and differences inside the same period, as well as phase alignment problems caused by multi-view sequences, we believe that it is a difficult and important task to adaptively encode the temporal periodic relationship.
To better exploit such periodic nature, we propose a plug-in module termed Temporal Periodic Alignment (TPA) in this paper. Key to our methods, TPA enhances the capacity of the transformer modules in exploiting temporal periodicity with two components, the first one is the Adaptive Fourier-transform Position Encoding (AFPE) which injects the periodic prior with adaptively superimposed position encodings based on Fourier-transform to generate the discriminative period-aware embeddings. Second, the period-aware embeddings are fed forward into the Temporal Aggregation Module (TAM) with explicit feature frequency decomposition and aggregation to explore the fine-grained dependencies.

Empirically, existing sequence-based baselines exhibit a performance enhancement when equipped with TPA strategy. We take a further step and establish a simple yet strong baseline named GaitFormer. We conduct extensive experiments on different benchmarks to demonstrate the effectiveness of our proposed method. In summary, our contributions are as follows:

\begin{itemize}
    \item{
    We propose Adaptive Fourier-transform Position Encoding (AFPE) module to force periodic prior for gait recognition. To the best of our knowledge, this is the first work to take full advantage of temporal periodicity. 
    } 
    \item{
    We present a plug-and-play strategy named Temporal Periodic Alignment (TPA), consisting of the novel AFPE and Temporal Aggregation Module (TAM). We further build a simple yet strong baseline GaitFormer based on the TPA module.
    }
    \item{
    By providing comprehensive experimental results, we show that the TPA module boosts the performance of existing methods as a plug-in complement. Moreover, our new proposed baseline GaitFormer outperforms SOTA methods on OU-MVLP (91.6\%), CASIA-B (98.9\%, normal) and GREW (64.8\%) datasets. 
    }
\end{itemize}

\section{Related Work}
\label{sec:related_work}

\paragraph{Gait Recognition.} Recent works for gait recognition can be mainly categorized into three types: template-based, image-based, and sequence-based approaches.

Early template-based methods directly employ the DNN on single gait templates to extract discriminate representations \cite{DBLP:conf/icb/ShiragaMMEY16,DBLP:conf/icassp/ZhangLMF16,DBLP:journals/pr/LiaoYAH20}. Gait Energy Image (GEI) \cite{DBLP:journals/pami/HanB06} is proposed as the temporal compression of a silhouette sequence for following template matching. Different architectures for CNNs are integrated \cite{DBLP:journals/pami/WuHWWT17} to facilitate cross-view gait recognition. Despite the simplicity of the template-based method, temporal and partial information might be missing during the template compression and the extracted representations tend to be sensitive to occlusion changes.

For the image-based strand \cite{DBLP:conf/accv/LiMXYYR20,DBLP:conf/cvpr/ChaiLZLW22}, GaitSet \cite{DBLP:conf/aaai/ChaoHZF19} incorporates horizontal segmentation and pyramid mapping to extract multi-scale spatial representations and aggregate them into a set-level feature. GaitPart \cite{DBLP:conf/cvpr/FanPC0HCHLH20} adopts the Focal Convolution Layer to enhance the part-level spatial features. GLN \cite{DBLP:conf/eccv/HouCLH20} leveraged the inherent feature pyramid to obtain more abundant spatial information. However, these methods neglect the motion changes and lost the dependencies of each part at different times during walking.

The sequence-based approaches \cite{DBLP:conf/eccv/DouZSYL22,DBLP:conf/eccv/LiangFHSHY22} regard the silhouettes of a gait sequence as a video for the comprehensive investigation of motion. Early methods \cite{DBLP:conf/icpr/FengLL16,DBLP:conf/ACISicis/NguyenHM18} adopt LSTM to extract the temporal relationship within segmented partitions. 

Xing et al. \cite{viewgait} , Zhen et al. \cite{huang20213d} 
 and GaitGL \cite{DBLP:conf/iccv/LinZ021} proposed 3D-CNN to capture spatial and temporal information simultaneously. TransGait \cite{li2022transgait} proposed Set Transformer to fuse sequence information while regarding gait sequences as sets. MetaGait \cite{dou2022metagait} learns omni-sample adaptive representation on spatial, temporal and channel dimensions.
Despite the success, these methods mostly integrate temporal representations with relatively simple paradigms and underemphasize the periodic inductive bias of gait recognition, whose capability is severely underestimated. In this paper, we advocate a temporal periodic alignment strategy to learn the periodic prior.

\begin{figure*}[t!] 
\centering 
\includegraphics[width=0.9\linewidth]{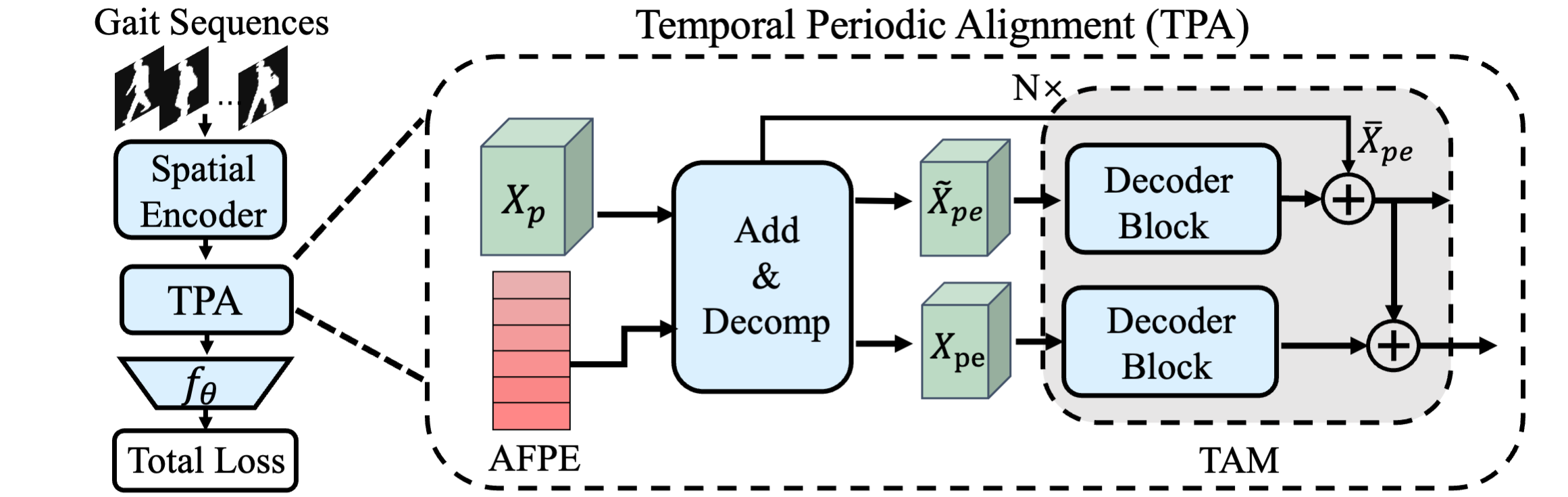}
\caption{Overview of the whole framework of our proposed GaitFormer. The Spatial Encoder generates the spatial feature $X_p$. Next, the AFPE encodes the discrete-time signals, added with $X_p$, generating period-aware embedding. Then, the embedding is decomposed into trend $\bar{X}_{pe}$ and seasonal part $\tilde{X}_{pe}$. They are sent to $N$ TAM blocks to learn temporal correlation information. Finally, a header $f_{\theta}$, consisting of BN and linear perception, is applied to map the feature vectors to the classification and metric space. Only the feature for metric learning is retained in the test stage.} 
\label{fig:framework} 
\end{figure*}

\paragraph{Periodicity in Gait Analysis.} Cunado et al. \cite{cunado1997using} applied Fourier Transform on two angle sequences from legs of model-based gait sequences to extract temporal features. An et al. \cite{an2020performance} utilized temporal normalize to sample a complete period from the gait sequences, which is different from our random sampling strategy to adaptively learn the periodic information. Li et al. \cite{li2022multi} exploit phase estimation method to explicitly synchronize the gait sequences under different views. On the contrary, we implicitly encode the sequences of different phases through AFPE to obtain phase-agnostic information extraction.

\paragraph{Transformer and Positional Encoding. } Transformer \cite{vaswani2017attention} has achieved remarkable achievements in natural language processing and computer vision. Vision Transformer \cite{dosovitskiy2020image} and its variants \cite{liu2021swin,liu2022video} have shown to excel and led to remarkable progress in visual recognition. Given the advantages in capturing correlations within long sequences, it is reasonable to apply this architecture to gait recognition. 

Positional encoding encrypts the positional correlation of the input sequence, which is crucial in Transformer-based long-sequence learning. Besides the vanilla positional encoding, several meticulously designed variants have recently been proposed \cite{chu2021conditional,liu2022petr} for different tasks. 

In this paper, based on the periodicity of gait recognition, we make the first attempt to apply the Fourier-transform to generate a series of base sequences and adaptively aggregate them as positional encoding, simulating periodic posture changes of pedestrians walking.

\section{Method}
In this section, we first define the formulation and outline the framework of our method. Then we elaborate on the temporal periodic alignment (TPA) strategy, which encapsulates the Adaptive Fourier-transform Position Encoding (AFPE) and Temporal Aggregation Module (TAM). Finally, we describe the details of the GaitFormer which is a simple yet strong baseline established on the basis of TPA.

\subsection{Architecture Overview}

The overview of our proposed pipeline is shown in Figure~\ref{fig:framework}. 
The Spatial Encoder comprises a CNN backbone and HP module, which is a common setting in GaitPart and OpenGait \cite{Fan_2023_CVPR}.
For a gait sequence from source dataset, the input frames are sampled as $ X_{i} \in \mathbb{R}^{C_{i} \times T_{i} \times H_{i} \times W_{i}} $, 
where $C_{i}$ denotes the channel number of the silhouette, $T_{i}$ is the length of the sampled sequence and $(H_{i}, W_{i})$ represents the image shape. The backbone generates output feature maps as $X_{o} \in \mathbb{R}^{C_{o} \times T_{o} \times H_{o}  \times W_{o} }$, where $C_{o}$ is output channel of the backbone. Since no downsampling occurs in the temporal dimension, $T_{o}$ is same as $T_{i}$, while the spatial shape of silhouettes undergoes downsampling twice.
Next, the Horizontal Pooling (HP) \cite{DBLP:conf/aaai/ChaoHZF19} module is applied to extract discriminative part-informed features $X_{p} \in \mathbb{R}^{P \times C_{o} \times T_{o}}$, where $P$ represents the number of horizontally split parts. 

Then, the part-level features are sent into TPA strategy. Specifically, the features are firstly fused with AFPE, followed by several blocks of TAM to learn the temporal correlation information within each sequence. Therefore, the output feature of TPA is denoted as $ X_{t} \in \mathbb{R}^{P  \times C_{t}} $, where $C_{t}$ is the channel of the output.

Finally, we use several separate BN \cite{ioffe2015batch} and linear perception layers, represented as $f_{\theta}$, to map the feature vectors to the classification and metric space for gait recognition.

\subsection{Temporal Periodic Alignment}
As aforementioned, previous works on gait recognition mostly focus on extracting better spatial representations, while underemphasizing the periodic dependencies along the temporal dimension. To overcome this weakness, we present a plug-and-play strategy termed Temporal Periodic Alignment (TPA) to force the periodic prior and explore fine-grained temporal dependencies, which comprise two key components AFPE and TAM in what follows.

\subsubsection{Adaptive Fourier-transform Position Encoding}
In order to better exploit the temporal correlation within the long gait sequence, we resort to the transformer network and design our TPA module based on the attention mechanism. Position encoding (PE) encodes the relative positional relationship in the transformer blocks. Considering that vanilla position encoding (PE) regards each element in the sequence as unique and thus fails to capture the periodic correlation variations, we aim to inject the periodic inductive bias in the position encoding. Based on the observations on differences in periods between different people, fine-grained similarity within period and phase align problems, 
we incorporate a Fourier transform to generate a series of base position encodings with different frequencies and rely on a position encoder to adaptively aggregate the final position encoding to represent ever-changing periodic information.

Specifically, the Discrete Fourier transform (DFT) is a mathematical transform that decomposes functions depending on space or time into its discrete frequencies, and the Inverse Discrete Fourier Transform (IDFT) is the inverse of DFT. According to Euler's formula, $e^{- i \theta } = cos(\theta ) - i \cdot sin(\theta )$, given an origin sequence $\{x_{t}\}$, containing $T$ elements, and a set of base sequences $\{X_{k}\}$, where $ k \in [0, T - 1] $, the IDFT is defined by the formula:
\begin{equation}  
\label{idft}
\begin{aligned}
 x_{t} & = \frac{1}{T}\sum_{k=1}^{T-1} X_{k} \cdot e^{i\frac{2 \pi}{T} k t }, \ \ \ \ \ \ \ \ \ \  t \in [0, T - 1] \\
 & =\frac{1}{T}\sum_{k=1}^{T-1} X_{k} \cdot (cos(\frac{2 \pi}{T} k t ) + i \cdot sin(\frac{2 \pi}{T} k t))
\end{aligned}
\end{equation}

Inspired by the above principles, as well as the idea of PETR \cite{liu2022petr}, in which the real physical location information in the image is transformed into the position encoding, we propose AFPE to encode a series of discrete-time signals into the position encoding and extend the original embeddings into period-aware ones.

Specifically, for a sequence of length T, we sample $T_d$ base sequences evenly from all $T$ components of IDFT tokens with interval $ d = \left \lfloor \frac{T}{T_d}  \right \rfloor $ to reduce computation and complexity. 
Consequently, the downsampled set of base sequences $\{\hat{X}_k\}, k = \{d, 2 \cdot d, ..., T_d \cdot d \} $ can be formulated:

\begin{equation}
\label{fpe_base}
\{\hat{X}_k\} = \{cos(\frac{2 \pi}{T} k t ), sin(\frac{2 \pi}{T} k t)\}, t \in [0, T - 1]
\end{equation}
Then, the position encoding procession can be formulated as follows:
\begin{equation}
\label{pe_mlp}
AFPE = \psi (Concat(\{\hat{X}_k\}))
\end{equation}
where $Concat(\cdot)$ means stack all the base sequences on the channel dimension, $\psi$ is the position encoding function, consists of a 2-layer Multi-Layer Perceptron (MLP), and the output shape is $AFPE \in \mathbb{R}^{T_i \times C_o}$. 

\begin{figure}[t!] 
\centering 
\includegraphics[width=0.95\linewidth]{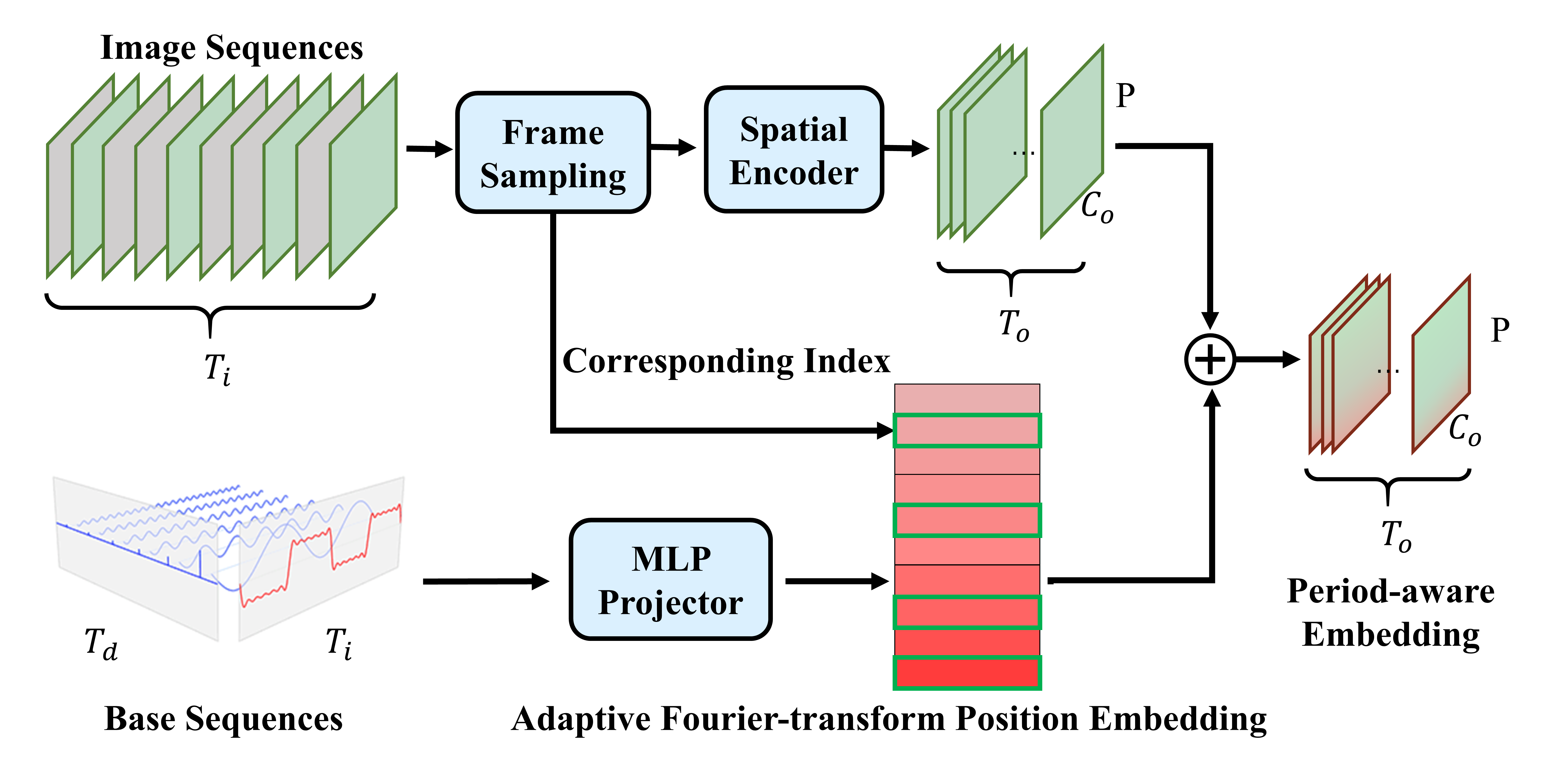} 
\caption{Illustration of the proposed Adaptive Fourier-transform Position Encoding (AFPE). The base sequences are sampled evenly from all DFT components. Then the bases are transformed into AFPE by a multi-layer perception. The corresponding index of the input frames is used to gather the relative AFPE. Finally, the AFPE is added with the features, producing period-aware embeddings. } 
\label{fig:fpe} 
\end{figure}

Meanwhile, the indexes of the input frames are used to gather corresponding features from the AFPE, which aims at encoding the relative temporal correlation. The abovementioned process can be expressed by the formula:
\begin{equation}  
\label{pe}
X_{pe} = X_{p} + gather(AFPE, indexes)
\end{equation}
where $X_{p}$ indicates the embedding feature of the HP module and $X_{pe}$ is the period-aware embedding, $indexes$ is the relative frame indexes of sampled input frames, $gather(\cdot)$ means sample the AFPE elements according to the indexes, as shown in Figure~\ref{fig:fpe}.

\begin{table*}[!t]
\begin{center}
\footnotesize
\centering
\renewcommand\arraystretch{1.1}

\begin{tabular}{lccccccccccccccc}
\hline
\multicolumn{1}{l|}{\multirow{2}{*}{{Methods}}} & \multicolumn{14}{c|}{{Probe View}}&\multicolumn{1}{|c}{\multirow{2}{*}{{Mean}}} \\ 
\cline{2-15} 
\multicolumn{1}{c|}{} &

\multicolumn{1}{c}{0$^{\circ}$}    &      \multicolumn{1}{c}{15$^{\circ}$}   & \multicolumn{1}{c}{30$^{\circ}$}   & \multicolumn{1}{c}{45$^{\circ}$}   & \multicolumn{1}{c}{60$^{\circ}$}   & \multicolumn{1}{c}{75$^{\circ}$}   & \multicolumn{1}{c}{90$^{\circ}$}   & \multicolumn{1}{c}{180$^{\circ}$}  & \multicolumn{1}{c}{195$^{\circ}$}  & \multicolumn{1}{c}{210$^{\circ}$}  & \multicolumn{1}{c}{225$^{\circ}$}  & \multicolumn{1}{c}{240$^{\circ}$}  & \multicolumn{1}{c}{255$^{\circ}$}  & \multicolumn{1}{c}{270$^{\circ}$}  & \multicolumn{1}{|c}{} \\ 
\toprule
\multicolumn{1}{l|}{GaitSet \cite{DBLP:conf/aaai/ChaoHZF19}} & \multicolumn{1}{c}{79.3} & \multicolumn{1}{c}{87.9} & \multicolumn{1}{c}{90.0} & \multicolumn{1}{c}{90.1} & \multicolumn{1}{c}{88.0} & \multicolumn{1}{c}{88.7}       & \multicolumn{1}{c}{87.7} & \multicolumn{1}{c}{81.8} & \multicolumn{1}{c}{86.5} & \multicolumn{1}{c}{89.0} & \multicolumn{1}{c}{89.2} & \multicolumn{1}{c}{87.2}       & \multicolumn{1}{c}{87.6} & \multicolumn{1}{c}{86.2} &  \multicolumn{1}{|c}{87.1} \\ 
\multicolumn{1}{l|}{GaitPart \cite{DBLP:conf/cvpr/FanPC0HCHLH20}} & \multicolumn{1}{c}{82.6} & \multicolumn{1}{c}{88.9} & \multicolumn{1}{c}{{90.8}} & \multicolumn{1}{c}{{91.0}}       & \multicolumn{1}{c}{89.7} & \multicolumn{1}{c}{89.9} & \multicolumn{1}{c}{89.5} & \multicolumn{1}{c}{85.2} & \multicolumn{1}{c}{{88.1}} & \multicolumn{1}{c}{{90.0}}       & \multicolumn{1}{c}{{90.1}} & \multicolumn{1}{c}{{89.0}} & \multicolumn{1}{c}{{89.1}} & \multicolumn{1}{c}{{88.2}} &  \multicolumn{1}{|c}{88.7} \\ 
\multicolumn{1}{l|}{GLN \cite{DBLP:conf/eccv/HouCLH20}} & \multicolumn{1}{c}{{83.8}}          & \multicolumn{1}{c}{{90.0}} & \multicolumn{1}{c}{{91.0}} & \multicolumn{1}{c}{{91.2}}       & \multicolumn{1}{c}{{90.3}} & \multicolumn{1}{c}{{90.0}} & \multicolumn{1}{c}{{89.4}}       & \multicolumn{1}{c}{{85.3}} & \multicolumn{1}{c}{89.1} & \multicolumn{1}{c}{90.5}       & \multicolumn{1}{c}{90.6} & \multicolumn{1}{c}{89.6} & \multicolumn{1}{c}{89.3}       & \multicolumn{1}{c}{{88.5}} &  \multicolumn{1}{|c}{89.2} \\ 

\multicolumn{1}{l|}{GaitGL \cite{DBLP:conf/iccv/LinZ021}} & \multicolumn{1}{c}{{84.9}}          & \multicolumn{1}{c}{{90.2}} & \multicolumn{1}{c}{{91.1}} & \multicolumn{1}{c}{{91.5}}       & \multicolumn{1}{c}{{91.1}} & \multicolumn{1}{c}{{90.8}} & \multicolumn{1}{c}{{90.3}}       & \multicolumn{1}{c}{{88.5}} & \multicolumn{1}{c}{88.6} & \multicolumn{1}{c}{90.3}       & \multicolumn{1}{c}{90.4} & \multicolumn{1}{c}{89.6} & \multicolumn{1}{c}{89.5}       & \multicolumn{1}{c}{{88.8}} &  \multicolumn{1}{|c}{89.7} \\ 

\multicolumn{1}{l|}{ReverseMask \cite{shen2022gait}} & \multicolumn{1}{c}{{87.9}}       & \multicolumn{1}{c}{{91.5}}       & \multicolumn{1}{c}{{91.7}}       & \multicolumn{1}{c}{{92.0}}       & \multicolumn{1}{c}{{92.0}}       & \multicolumn{1}{c}{{91.6}}       & \multicolumn{1}{c}{{91.3}}       & \multicolumn{1}{c}{{90.7}}       & \multicolumn{1}{c}{{90.3}}       & \multicolumn{1}{c}{{90.9}}       & \multicolumn{1}{c}{{91.1}}       & \multicolumn{1}{c}{{90.8}}       & \multicolumn{1}{c}{{90.5}}       & \multicolumn{1}{c}{{90.2}}       &  \multicolumn{1}{|c}{{90.9}} \\ 

\multicolumn{1}{l|}{3DLocal \cite{huang20213d}} & \multicolumn{1}{c}{{86.1}}       & \multicolumn{1}{c}{{91.2}}       & \multicolumn{1}{c}{{92.6}}       & \multicolumn{1}{c}{{92.9}}       & \multicolumn{1}{c}{{92.2}}       & \multicolumn{1}{c}{{91.3}}       & \multicolumn{1}{c}{{91.1}}       & \multicolumn{1}{c}{{86.9}}       & \multicolumn{1}{c}{{90.8}}       & \multicolumn{1}{c}{\textbf{92.2}}       & \multicolumn{1}{c}{\textbf{92.3}}       & \multicolumn{1}{c}{{91.3}}       & \multicolumn{1}{c}{{91.1}}       & \multicolumn{1}{c}{{90.2}}       &  \multicolumn{1}{|c}{{90.9}} \\ 

\multicolumn{1}{l|}{\textbf{GaitFormer}} & \multicolumn{1}{c}{\textbf{89.6}}       & \multicolumn{1}{c}{\textbf{92.3}}       & \multicolumn{1}{c}{\textbf{92.1}}       & \multicolumn{1}{c}{\textbf{92.4}}       & \multicolumn{1}{c}{\textbf{92.5}}       &
\multicolumn{1}{c}{\textbf{91.9}}       & \multicolumn{1}{c}{\textbf{91.7}}       & \multicolumn{1}{c}{\textbf{91.5}}       & \multicolumn{1}{c}{\textbf{91.6}}       & \multicolumn{1}{c}{{91.5}}       & \multicolumn{1}{c}{{91.7}}       & \multicolumn{1}{c}{\textbf{91.7}}       & \multicolumn{1}{c}{\textbf{91.2}}       & \multicolumn{1}{c}{\textbf{91.0}}       & \multicolumn{1}{|c}{\textbf{91.6}} \\


\midrule
\midrule
\multicolumn{1}{l|}{GaitSet \cite{DBLP:conf/aaai/ChaoHZF19}} & \multicolumn{1}{c}{84.5} & \multicolumn{1}{c}{93.3} & \multicolumn{1}{c}{96.7} & \multicolumn{1}{c}{96.6} & \multicolumn{1}{c}{93.5} & \multicolumn{1}{c}{95.3}       & \multicolumn{1}{c}{94.2} & \multicolumn{1}{c}{87.0} & \multicolumn{1}{c}{92.5} & \multicolumn{1}{c}{96.0} & \multicolumn{1}{c}{96.0} & \multicolumn{1}{c}{93.0}       & \multicolumn{1}{c}{94.3} & \multicolumn{1}{c}{92.7} &  \multicolumn{1}{|c}{93.3} \\ 
\multicolumn{1}{l|}{GaitPart \cite{DBLP:conf/cvpr/FanPC0HCHLH20}} & \multicolumn{1}{c}{88.0} & \multicolumn{1}{c}{94.7} & \multicolumn{1}{c}{{97.7}} & \multicolumn{1}{c}{{97.6}}       & \multicolumn{1}{c}{95.5} & \multicolumn{1}{c}{96.6} & \multicolumn{1}{c}{96.2} & \multicolumn{1}{c}{90.6} & \multicolumn{1}{c}{{94.2}} & \multicolumn{1}{c}{{97.2}}       & \multicolumn{1}{c}{{97.1}} & \multicolumn{1}{c}{{95.1}} & \multicolumn{1}{c}{{96.0}} & \multicolumn{1}{c}{{95.0}} &  \multicolumn{1}{|c}{95.1} \\ 
\multicolumn{1}{l|}{GLN \cite{DBLP:conf/eccv/HouCLH20}} & \multicolumn{1}{c}{{89.3}}          & \multicolumn{1}{c}{{95.8}} & \multicolumn{1}{c}{{97.9}} & \multicolumn{1}{c}{{97.8}}       & \multicolumn{1}{c}{{96.0}} & \multicolumn{1}{c}{{96.7}} & \multicolumn{1}{c}{{96.1}}       & \multicolumn{1}{c}{{90.7}} & \multicolumn{1}{c}{95.3} & \multicolumn{1}{c}{97.7}       & \multicolumn{1}{c}{97.5} & \multicolumn{1}{c}{95.7} & \multicolumn{1}{c}{96.2}       & \multicolumn{1}{c}{{95.3}} &  \multicolumn{1}{|c}{95.6} \\ 

\multicolumn{1}{l|}{GaitGL \cite{DBLP:conf/iccv/LinZ021}} & \multicolumn{1}{c}{{90.5}}          & \multicolumn{1}{c}{{96.1}} & \multicolumn{1}{c}{{98.0}} & \multicolumn{1}{c}{{98.1}}       & \multicolumn{1}{c}{{97.0}} & \multicolumn{1}{c}{{97.6}} & \multicolumn{1}{c}{{97.1}}       & \multicolumn{1}{c}{{94.2}} & \multicolumn{1}{c}{94.9} & \multicolumn{1}{c}{97.4}       &
\multicolumn{1}{c}{97.4}       & \multicolumn{1}{c}{95.7} & \multicolumn{1}{c}{96.5} & \multicolumn{1}{c}{95.7}       &  \multicolumn{1}{|c}{96.2} \\ 
\multicolumn{1}{l|}{ReverseMask \cite{shen2022gait}} & \multicolumn{1}{c}{{93.7}}       & \multicolumn{1}{c}{{97.5}}       & \multicolumn{1}{c}{{98.6}}       & \multicolumn{1}{c}{{98.8}}       & \multicolumn{1}{c}{{98.0}}       & \multicolumn{1}{c}{{98.5}}       & \multicolumn{1}{c}{{98.2}}       & \multicolumn{1}{c}{{96.5}}       & \multicolumn{1}{c}{{96.7}}       & \multicolumn{1}{c}{{98.2}}       & \multicolumn{1}{c}{{98.1}}       & \multicolumn{1}{c}{{97.1}}       & \multicolumn{1}{c}{{97.6}}       & \multicolumn{1}{c}{{97.2}}       &  \multicolumn{1}{|c}{{97.5}} \\ 

\multicolumn{1}{l|}{\textbf{GaitFormer}} & \multicolumn{1}{c}{\textbf{95.4}}       & \multicolumn{1}{c}{\textbf{98.3}}       & \multicolumn{1}{c}{\textbf{99.0}}       & \multicolumn{1}{c}{\textbf{99.2}}       & \multicolumn{1}{c}{\textbf{98.5}}       & \multicolumn{1}{c}{\textbf{98.8}}       & \multicolumn{1}{c}{\textbf{98.6}}       & \multicolumn{1}{c}{\textbf{97.4}}       & \multicolumn{1}{c}{\textbf{98.0}}       & \multicolumn{1}{c}{\textbf{98.8}}       & \multicolumn{1}{c}{\textbf{98.7}}       & \multicolumn{1}{c}{\textbf{98.1}}       & \multicolumn{1}{c}{\textbf{98.4}}       &   \multicolumn{1}{c}{\textbf{98.1}}       & \multicolumn{1}{|c}{\textbf{98.2}} \\ 


\bottomrule

\end{tabular}
\caption{Rank-1 accuracy (\%) of the compared method on OU-MVLP under 14 probe views excluding identical-view cases. The top 7 rows and bottom 6 rows show the results
with and without invalid probe sequences respectively.}
\label{tab:oumvlp_results}
\end{center}
\end{table*}

\subsubsection{Temporal Aggregation Module}

After embeddings of a sequence are injected with periodic prior through AFPE, the fine-grained temporal dependencies remain underexplored. In pursuit of grabbing features that are more related to the walking pattern and robust to view change, We revamp the vanilla decoder of transformer from the perspective of trend-seasonal decomposition. 

Specifically, different from the traditional decoder module, the Decoder block of Temporal Periodic Module is composed of  Multi-Head Cross-Attention (MHCA) \cite{vaswani2017attention} sub-layer and a position-wise fully connected feed-forward network (FFN), where the period-aware embeddings and class-tokens are directly sent into MHCA sub-layer to extract temporal information. 
In order to learn periodic temporal information, we decompose the inputting features into the trend-cyclical and the seasonal part, according to the common practice of time series forecasting problems \cite{cleveland1990stl}. Here, we set trend-cyclical as the average mean of a single gait period, which is about 30 frames according to statistical analysis. After that, origin features and seasonal features are sent to their respective transformer modules with corresponding tokens, then the output tokens are fused through linear lateral layers. TAM explicitly learns the frequency decomposition of features to balance temporal sensitivity and robustness.
The procession can be described as follow:

\begin{equation}  
\label{tpm}
\begin{aligned}
& \hat{x}_{token}^{i} = FFN(MHCA(x_{token}^{i-1}, X_{pe}))
 \\
& \tilde{x}_{token}^{i} = FFN(MHCA(\tilde{x}_{token}^{i-1}, \tilde{X}_{pe}))
 \\
& x_{token}^{i} = Fusion(\hat{x}_{token}^{i}, Lateral(\bar{X}_{pe} + \tilde{x}_{token}^{i}))
\end{aligned}
\end{equation}
where 
$\bar{X}_{pe}$ and $\tilde{X}_{pe}$ are the calculated trend-cyclical and seasonal part of the embedding, $x_{token}^{i-1}$ and $x_{token}^{i}$ indicate the token for full feature of previous transformer block and current $i th$ block. $\hat{x}_{token}^{i}$ represents the token for residual fusion, while $\tilde{x}_{token}^{i}$ indicates the token for seasonal feature. The $Lateral(\cdot)$ function can be one MLP layer, meanwhile the $Fusion(\cdot)$ function is $Mean(\cdot)$ in practice. Note that descriptions of operations like Layer-Norm and residual connection are omitted, which means they are the same as common practices.

\subsection{Training objectives}
In our work, we incorporate a combined loss, consist of triplet loss \cite{hoffer2015deep} and cross-entropy loss to effectively train the proposed gait recognition model.  The temporal triplet loss aims to improve the inter-class distance and reduce the intra-class distance, which is more beneficial to distinguishing untrained human IDs. In our work, we adapt triplet loss with temporal dimension, to cope with the distance comparison between the temporal features.
\begin{equation}
\label{triplet_loss}
\begin{aligned}
L_{tri} = [ &Dist(\mathcal{F}(X_{a} ), \mathcal{F}(X_{p})) -
\\
&Dist(\mathcal{F}(X_{a}), \mathcal{F}(X_{n})) + m]_{+}
\end{aligned}
\end{equation}
where $X_{a}$ and $X_{p}$ are samples from the same class, $X_{n}$ is from another class, $\mathcal{F}$ represents our proposed feature extractor and mapping method. $m$ is the margin of the triplet loss. The operation $[\cdot]_{+}$ is equal to $Max(\cdot, 0)$.

\begin{equation}
\label{temporal_dist}
Dist(X, Y) = \frac{1}{T}\sum^{T-1}_{t=0} \sqrt{(x_t - y_t)^2}
\end{equation}
Specifically, $Dist(\cdot)$ is the mean of Euclidean distances between $x_t$ and $y_t$ along temporal dimension. 

Further, we replace the vanilla cross-entropy loss with ArcFace\cite{deng2019arcface} loss to encourage a larger margin, which is denoted as $L_{cls}$. The combined loss $L$ can be defined as:
\begin{equation}
\label{total_loss}
L = p * L_{cls} + q * L_{tri}
\end{equation}
where $p$ and $q$ are the weight of losses. The hyper-parameters in Equation \ref{total_loss} are described in Implementation Details.

\section{Experiments}
Our empirical experiments mainly consist of two parts. In the first part, extensive experiments are conducted on three popular gait datasets: CASIA-B \cite{DBLP:conf/icpr/TanHYT06}, OU-MVLP \cite{DBLP:journals/ipsjtcva/TakemuraMMEY18} and GREW \cite{zhu2021gait}. Then ablation experiments and analysis of the performance are given in the second part.

\subsection{Datasets}
\paragraph{CASIA-B} dataset is the most popular gait dataset, including RGB images and silhouettes of 124 subjects with 10 sequences under 11 views ($0^\circ$,$18^\circ$, ......,$180^\circ$). The 10 sequences contain 6 normal walking variants (NM), 2 variants with subjects carrying bags (BG), and 2 variants with subjects wearing different clothes (CL). Hence, each subject contains 10 (groups) × 11(view angle) = 110 gait sequences. We strictly follow the popular protocol for evaluation, which employs the first 74 subjects as the training set and the latter 50 subjects are reserved
for testing. 

\paragraph{OU-MVLP} dataset is a larger public gait dataset created by Osaka University, which includes 10,307 subjects. Each subject was collected under 14 views between [$0^\circ$, $90^\circ$] and [$180^\circ$,$270^\circ$] with the interval of $15^\circ$, and each view consist of 2 video sequences (Seq\#00 and Seq\#01). During testing, the sequences in Seq\#01 serve as the gallery set and those in Seq\#00 are regarded as the probe set.

\paragraph{GREW} dataset is the latest gait dataset which is more challenging. It consists of 26,345 subjects and 128,671 sequences, obtained with 882 cameras in wild environments. The GREW dataset has diverse and practical view variations, as well as more naturally challenging factors. Besides, the unconstrained setting also brings new challenging factors for gait patterns, such as diverse views, dressing, carrying, and crowd.

\begin{table}[t!]
  \centering
  \small
  \renewcommand\arraystretch{1.0} 
  
    \begin{tabular}{l|cccc}
    \toprule
    Methods & \multicolumn{1}{c}{Rank-1} & \multicolumn{1}{c}{Rank-5} & \multicolumn{1}{c}{Rank-10} & \multicolumn{1}{c}{Rank-20} \\
    \midrule
    GEINet \cite{shiraga2016geinet} & 6.8   & 13.4  & 17.0  & 21.0  \\

    TS-CNN \cite{wu2016comprehensive} & 13.6  & 24.6  & 30.2  & 37.0  \\

    GaitSet \cite{DBLP:conf/aaai/ChaoHZF19}  & 46.3  & 63.6  & 70.3  & 76.8  \\
 
    GaitPart \cite{DBLP:conf/cvpr/FanPC0HCHLH20}  & 44.0  & 60.7  & 67.3  & 73.5  \\
 
    CSTL \cite{huang2021context}  & 50.6  & 65.9  & 71.9  & 76.9  \\

    TransGait  & 56.27  & 72.72 & 78.12  & 82.51  \\ \midrule

    \textbf{GaitFormer} & \textbf{65.53} & \textbf{78.63} & \textbf{83.04} & \textbf{86.38} \\
    \bottomrule
    \end{tabular}%
    \caption{Rank-1 accuracy (\%), Rank-5 accuracy (\%), Rank-10 accuracy (\%), and Rank-20 accuracy (\%) on the GREW dataset.}

  \label{tab:grew_results}%
\end{table}%

\subsection{Implementaion Details} 

All models are implemented in PyTorch \cite{paszke2019pytorch}. The silhouettes in both datasets are pre-processed using methods proposed in \cite{8063344}. The input size is set to 64x44. In a mini-batch, the number of subjects and the number of sequences are set to (8, 16) for CASIA-B and (32, 8) for OU-MVLP. In the training phase, we randomly sample the silhouettes from each gait sequence according to TSN sampling \cite{wang2018temporal}, the number of sampled frames is fixed at 30. While in the test phase, the silhouettes are sampled at the center of every segment.

AdamW \cite{loshchilov2017decoupled} optimizer is used, while weight decay is set to 5e-1. The learning rate is initialized to 1e-3 and decreased to 1e-6 in cosine decay style within 100,000 iterations. Besides, the warmup strategy is adopted at the start of training. As for hyperparameter settings in losses, the scale of Arcface loss $s$ is set to 32 and the margin $m$ is 0.3. Label smoothing is used to avoid overfitting. The margin threshold $m$ for triplet loss is set to 0.2. The ratio of Equation \ref{total_loss} is set as $p = 0.1$ and $q = 1.0$.

 The widely employed Rank-1 accuracy excluding identical-view sequences is taken as the evaluation metric for performance comparison. For the evaluation of CASIA-B, experiments are conducted with the setting LT. 

\begin{table}[t!]
 \centering
 \small

\renewcommand\arraystretch{1.0}
\tabcolsep=0.3cm
 \begin{tabular}{lcccc}
    \toprule
     \multicolumn{1}{l|}{\multirow{2}{*}{{Methods}}} &  \multicolumn{3}{c|}{{{Probe}}}
     & \multicolumn{1}{c}{\multirow{2}{*}{{Mean}}}\\
     \cline{2-4}
     \multicolumn{1}{c|}{} &
 \multicolumn{1}{c}{{{NM}}} 
 &
 \multicolumn{1}{c}{{{BG}}} 
 &
 \multicolumn{1}{c|}{{{CL}}} \\

    \midrule
     \multicolumn{1}{l|}{GaitSet \cite{DBLP:conf/aaai/ChaoHZF19}} & \multicolumn{1}{c}{95.0}  &\multicolumn{1}{c}{87.2}  & \multicolumn{1}{c|}{70.4}  & 84.2
\\ 
     \multicolumn{1}{l|}{GaitPart \cite{DBLP:conf/cvpr/FanPC0HCHLH20}} & \multicolumn{1}{c}{96.2}  &\multicolumn{1}{c}{91.5}  & \multicolumn{1}{c|}{78.7}  & 88.8
\\
    \multicolumn{1}{l|}{ReverseMask \cite{shen2022gait}}   & \multicolumn{1}{c}{97.7}   & \multicolumn{1}{c}{\textbf{95.3}}   &  \multicolumn{1}{c|}{86.0} & 93.0 
\\
     \multicolumn{1}{l|}{GaitGL \cite{DBLP:conf/iccv/LinZ021}}   & \multicolumn{1}{c}{97.4}   & \multicolumn{1}{c}{94.5}  & \multicolumn{1}{c|}{83.6} & 91.8  
\\
    \multicolumn{1}{l|}{3DLocal \cite{huang20213d}}    & \multicolumn{1}{c}{97.5}  &
 \multicolumn{1}{c}{94.3} & \multicolumn{1}{c|}{83.7}
 & 91.8
\\
\multicolumn{1}{l|}{GaitFormer}    & \multicolumn{1}{c}{96.1}  &
 \multicolumn{1}{c}{90.1} & \multicolumn{1}{c|}{75.8}
 & 87.3
\\ \midrule
\multicolumn{1}{l|}{GaitSet+ \cite{DBLP:conf/aaai/ChaoHZF19}}    &\multicolumn{1}{c}{{{96.9}}}  & \multicolumn{1}{c}{{92.6}} & \multicolumn{1}{c|}{81.7} & 90.4
\\ 

\multicolumn{1}{l|}{GaitPart+ \cite{DBLP:conf/cvpr/FanPC0HCHLH20}}    &\multicolumn{1}{c}{{{97.7}}}  & \multicolumn{1}{c}{{93.1}} & \multicolumn{1}{c|}{81.2} & 90.7
\\ 

    \multicolumn{1}{l|}{GaitGL+ \cite{DBLP:conf/iccv/LinZ021}}    &\multicolumn{1}{c}{{{98.5}}}  & \multicolumn{1}{c}{\textbf{95.3}} & \multicolumn{1}{c|}{85.8} & 93.2
\\ 

 \multicolumn{1}{l|}{GaitFormer+}   &
 \multicolumn{1}{c}{{\textbf{98.9}}}& \multicolumn{1}{c}{{94.7}}  &  \multicolumn{1}{c|}{\textbf{87.3}} & \textbf{93.6}
\\
    \bottomrule

    \end{tabular}%
    \caption{Rank-1 accuracy (\%) of the compared method on CASIA-B under different conditions, excluding identical-view cases with the setting of large-scale training (LT). Here methods with '+' denote two-stage training, warm-up on OU-MVLP and continue the second stage of training and testing on CASIA-B dataset.}
    \label{tab:casiab_results}

\end{table}
\begin{table}[t!]
    \small
     \centering
     \renewcommand{\arraystretch}{1.1}
    \tabcolsep=0.35cm
     \begin{tabular}{l|c|c}
   \toprule
            Methods & TPA & Mean Accuracy\\
            \midrule
            \multirow{2}{*}[-0.3ex]{GaitPart \cite{DBLP:conf/cvpr/FanPC0HCHLH20}} & \xmark  & 88.7 \\
             &$\checkmark$  & 89.4 \\
             \midrule
            \multirow{2}{*}[-0.3ex]{GaitGL \cite{DBLP:conf/iccv/LinZ021}} & \xmark    & 89.7 \\
             & $\checkmark$  & 90.4 \\
    \bottomrule
     \end{tabular}
    \caption{\small Results of baselines with and without TPA strategy on OU-MVLP dataset with invalid probe sequences. }
    \label{tab:plug}%
\end{table}

\begin{figure*}[t!]
     \centering
      \begin{subfigure}[b]{0.23\linewidth}
         \centering
         \includegraphics[width=\columnwidth]{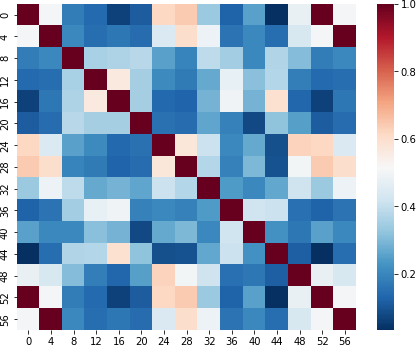}
         \caption{GaitPart feature.}
         \label{fig:heatmap_gaitpart}
     \end{subfigure}
      \begin{subfigure}[b]{0.23\linewidth}
         \centering
         \includegraphics[width=0.95\columnwidth]{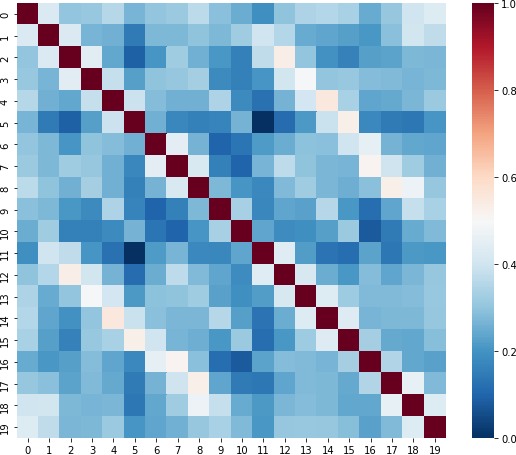}
         \caption{GaitGL feature.}
         \label{fig:heatmap_gaitgl}
     \end{subfigure}
    \begin{subfigure}[b]{0.24\linewidth}
       \centering
       \includegraphics[width=0.97\columnwidth]{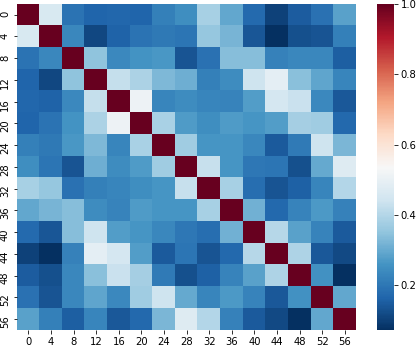}
       \caption{GaitFormer feature w/o TPA.}
       \label{fig:heatmap_backbone}
    \end{subfigure}
     \begin{subfigure}[b]{0.23\linewidth}
         \centering
         \includegraphics[width=\columnwidth]{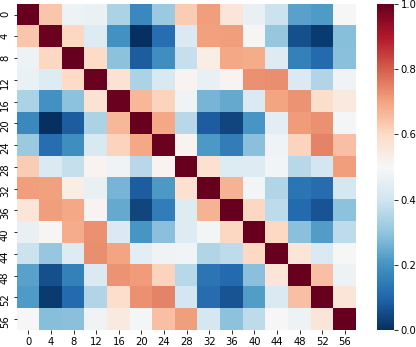}
         \caption{GaitFormer feature w/ TPA.}
         \label{fig:heatmap_gaitformer}
     \end{subfigure}
     
     \caption{Visualization of self-similarity of temporal features in gait recognition models. (a) shows that except for itself, the similarity between the features 30 frames away is relatively high as well, indicating that GaitPart \cite{DBLP:conf/cvpr/FanPC0HCHLH20} with little temporal aggregation can learn the periodic information somehow. Similarity in (b) shows that GaitGL \cite{DBLP:conf/iccv/LinZ021} learns better temporal periodic correlation.
     Correspondingly, (c) and (d) is the feature similarity of GaitFormer, with and without TPA respectively. It is obvious that the TPA module adaptively strengthens the periodic nature of temporal features. Note that, the horizontal and vertical coordinates in the heatmap represent frame indexes in the gait sequence, with red pixels indicating high similarity and blue the opposite. Due to 3D convolution with strides greater than 1, the frame number of (b) is less than others.}
     \label{fig:heatmap}
\end{figure*}

\begin{figure}[t!]
  \centering
  \includegraphics[width=\columnwidth]{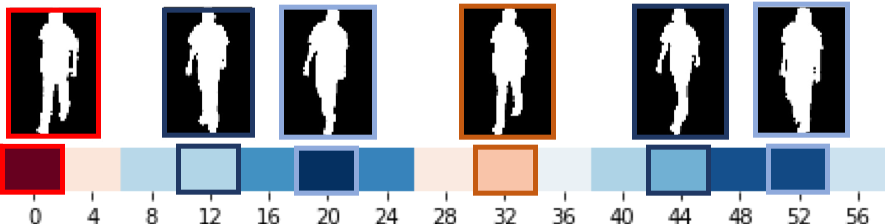}
   \caption{Visualization of silhouettes and feature similarity. The color bar is the first row of Figure~\ref{fig:heatmap_gaitformer}, in which red boxes mean high similarity and blue ones mean low. The silhouettes are sampled according to the corresponding indexes on the heatmap. We can observe that the silhouettes represented by features with higher similarity are more similar in appearance, and vice versa.}
   \label{fig:oumvlp_vis}
\end{figure}

\subsection{Comparison with State-of-the-Art Methods}

We first compare our method with several state-of-the-art methods including GaitSet \cite{DBLP:conf/aaai/ChaoHZF19}, GaitPart \cite{DBLP:conf/cvpr/FanPC0HCHLH20}, GaitGL \cite{DBLP:conf/iccv/LinZ021}, ReserveMask\cite{DBLP:journals/corr/abs-2203-04038},  TransGait \cite{li2022transgait} , CSTL \cite{huang2021context}, TS-CNN \cite{wu2016comprehensive} and 3DLocal \cite{huang20213d} on the OU-MVLP, CASIA-B, and GREW datasets. The results are listed as follows.

\subsubsection{Results on OU-MVLP}
As shown in Table~\ref{tab:oumvlp_results}, GaitFormer outperforms all its counterparts and achieves new SOTA results under various cross-view conditions on the OU-MVLP dataset. Specifically, the average accuracy of our method achieves 91.6\%, improved by 1.6\% and 0.7\% compared with advanced GaitGL and 3DLocal respectively. Besides, when the subjects in the probe without corresponding samples are discarded, GaitFormer consistently leads by an even larger margin. The results demonstrate the superiority of GaitFormer and indicate that GaitFormer has great potential for learning discriminative representations with more fine-grained temporal dependencies among cross-view walking silhouettes.

\subsubsection{Results on CASIA-B and GREW}
The results on CASIA-B dataset and GREW dataset are shown in Table~\ref{tab:casiab_results} and Table~\ref{tab:grew_results} respectively. GaitFormer surpasses other SOTA methods on both the small-scale CASIA-B dataset and large-scale GREW dataset. It is worth noting that the number of samples of CASIA-B is too limited and easily overfitted with the transformer framework, the original GaitFormer bears an observable performance drop. As a simple remedy, we utilize the model pre-trained on OU-MVLP to finetune on the CASIA-B dataset, which enables the GaitFormer with a wide leading margin. In the
challenging setup of GREW, GaitFormer still achieves promising results, rank-1 accuracy reaches 65.53\%, which is significantly higher than previous methods. The results validate the effectiveness of our framework, showing that GaitFormer is robust even in the presence of wild and unconstrained settings.

\begin{table}[t]
\centering
\small
\renewcommand\arraystretch{1.0} 

\begin{tabular}{l|ccc}
\toprule

Methods & FLOPs &Params & Accuracy \\

\midrule
GaitSet \cite{DBLP:conf/aaai/ChaoHZF19}	&12.91G & 6.3M & 87.1 \\

GaitPart \cite{DBLP:conf/cvpr/FanPC0HCHLH20} &7.93G & 4.8M & 88.7 \\

GLN \cite{DBLP:conf/eccv/HouCLH20} &73.55G & 8.5M & 89.2 \\

GaitGL \cite{DBLP:conf/iccv/LinZ021}	&58.64G & 95.6M & 89.7 \\

3DLocal \cite{huang20213d} & 11.17G & 4.3M & 90.9 \\
\midrule
GaitFormer 	&9.62G & 51.0M & \textbf{91.6}\\

\bottomrule
\end{tabular}
\caption{ The analysis of model complexity on OU-MVLP. The number of average floating-point operations (FLOPs) and the parameter number (Params) are reported.}
\label{tab_complexity}
\end{table}

\begin{table}[t!]
  \small
  \centering
 
  \renewcommand\arraystretch{1.0} 
     
    \begin{tabular}{l|cccc}
    \toprule
    Layer Number & \multicolumn{1}{c}{n = 4} & \multicolumn{1}{c}{n = 5} & \multicolumn{1}{c}{n = 6} & \multicolumn{1}{c}{n = 7} \\
    \midrule
    Rank-1 Accuracy  & 90.85 & 90.71 & \textbf{91.6} & 91.26 \\
    \bottomrule
    \end{tabular}%
  \caption{The ablation study for layer number in TAM. }

   \label{tab:tpm_ablation}%
\end{table}%

\begin{table*}[t!]
    \small
     \centering
     \renewcommand{\arraystretch}{1.1}
    \tabcolsep=0.35cm
     \begin{tabular}{l|ccc|cc}
   \toprule
            Ablation & Temporal Operation & AFPE & TAM &  OU-MVLP & GREW\\
            \midrule
            
            \textbf{GaitFormer} & Transformer & $\checkmark$  & $\checkmark$ & \textbf{91.60} & \textbf{65.53} \\ \midrule
            GaitFormer w/o TPA & Max Pooling  & \xmark  & \xmark &  87.20 & 50.94 \\
            GaitFormer with decoder & vanilla Transformer & \xmark  & \xmark &  90.51 & 62.52 \\
            GaitFormer w/o AFPE & Transformer & \xmark  & $\checkmark$   & 91.24 & 63.74\\
            GaitFormer w/o TAM & Transformer & $\checkmark$  & \xmark  & 91.20 & 63.50 \\
    \bottomrule
     \end{tabular}
    \caption{\small Ablation study of GaitFormer on OU-MVLP and GREW dataset. The number of transformer blocks is set as 6 in experiments. }
    \label{tab:ablation}%
\end{table*}

\subsection{Ablation Study}
Several ablation studies with various settings will be conducted on the OU-MVLP dataset to verify the effectiveness of each component in GaitFormer, including the components ablation, $T_d$ in AFPE, layer numbers in TAM, and the results of other baselines plugged with TPA.

\begin{figure}[t!]
     \centering
     \begin{subfigure}[b]{0.49\linewidth}
         \centering
         \includegraphics[width=\columnwidth]{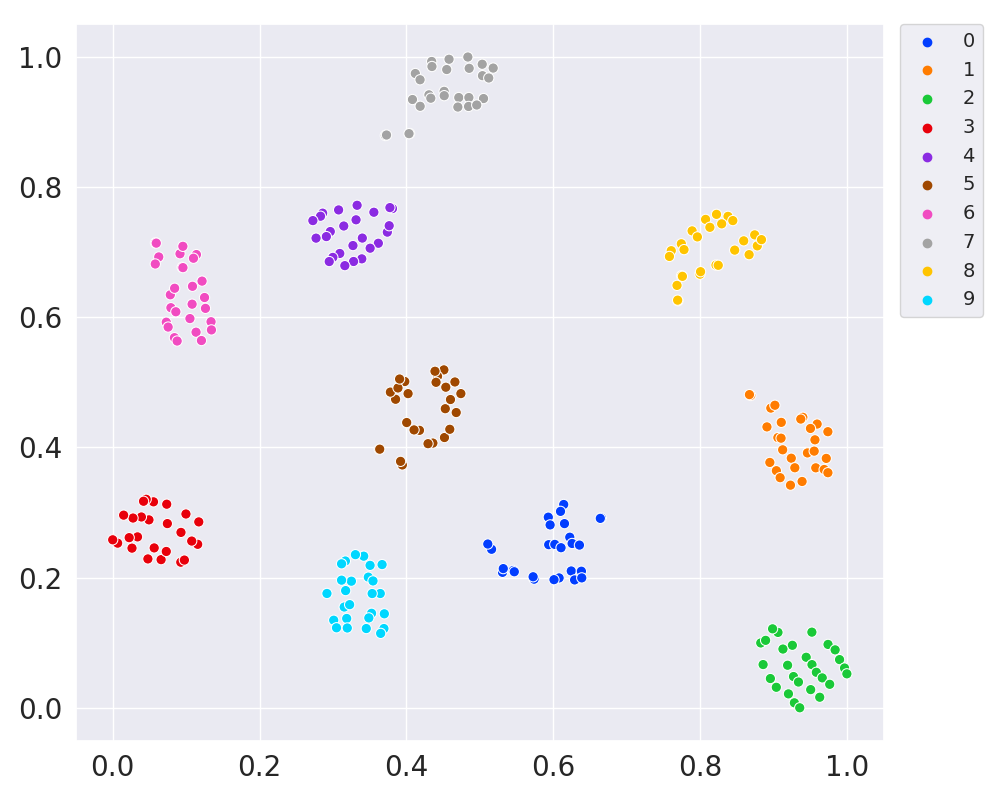}
         \caption{GaitFormer w/o TPA.}
         \label{fig:tsne_gaitformer_bad}
     \end{subfigure}
     \hfill
     \begin{subfigure}[b]{0.49\linewidth}
         \centering
         \includegraphics[width=\columnwidth]{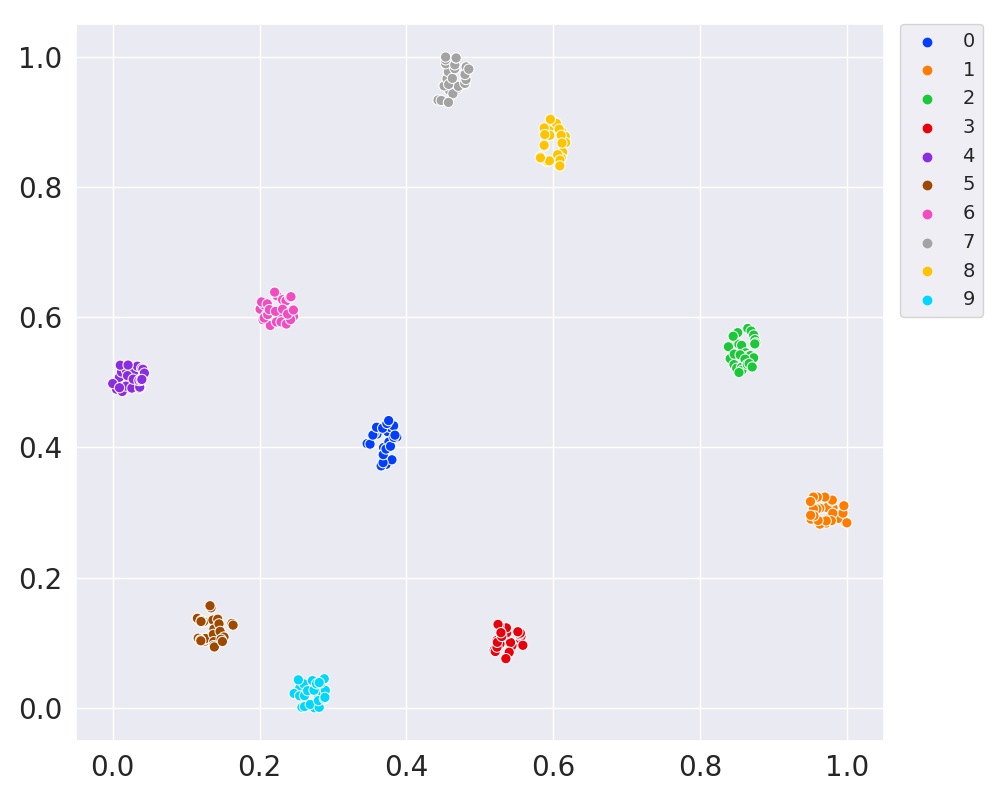}
         \caption{GaitFormer with TPA.}
         \label{fig:tsne_gaitformer}
     \end{subfigure}
     \caption{(a) and (b) show the t-SNE visualization examples of GaitFormer w/o TPA and GaitFormer on OU-MVLP test dataset. We visualize 10 identities and each point represents a sample and each color defines a class.}
     \label{fig:tsne}
\end{figure}

\paragraph{GaitFormer Components Ablation.}
To explore the contribution of the baseline and proposed modules, we conduct ablation experiments on the OU-MVLP dataset to exploit the role of different parts. The experimental results are shown in Table~\ref{tab:ablation}. We can observe that GaitFormer with little temporal pooling achieves poor performance, while the absence of AFPE or TAM brings about severe performance drop, which signifies the importance of these components for producing better representations for gait recognition. Figure~\ref{fig:tsne} consistently demonstrates the effectiveness of TPA from the perspective of feature visualization.

\paragraph{Results of other baselines plugged with TPA}
To verify that our proposed TPA module is an easy-to-use plug-in technique, we replace the temporal aggregation component of other sequence-based baselines with the TPA module. The results of these baselines plugged with the TPA module are shown in Table~\ref{tab:plug}. As we can see, the performance of these methods was further improved when integrated with TPA. This demonstrates that TPA consistently delivers excellent performance with different backbones, proving its portability and robustness.

\begin{figure}[t!]
     \centering
         \centering
         \includegraphics[width=0.66\columnwidth]{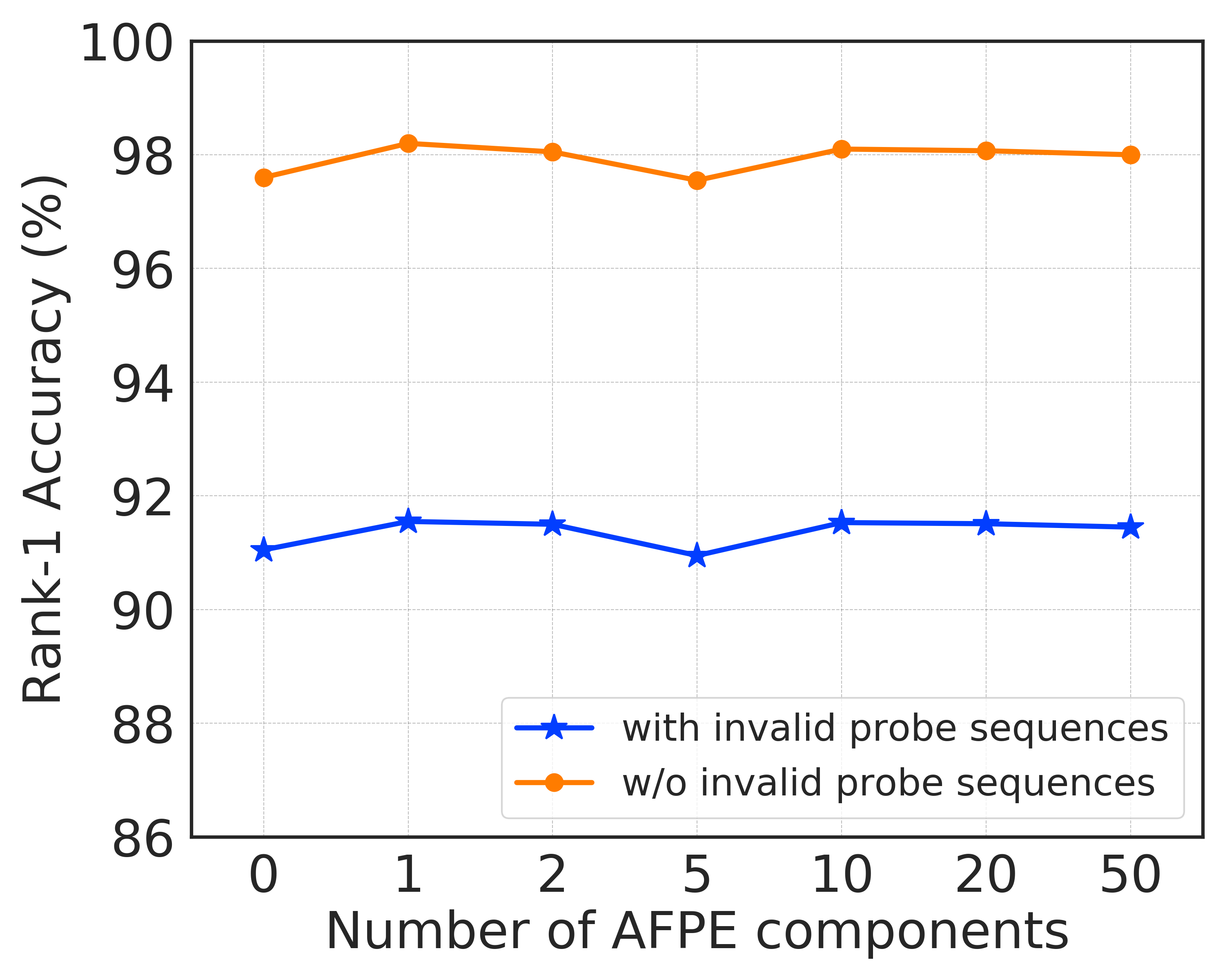}
     \caption{ The average rank-1 accuracies with different numbers of AFPE components on the OU-MVLP dataset.}
     \label{fig:ablation_FPE}
\end{figure}

\paragraph{GaitFormer taps into periodic dependencies}
To further prove that GaitFormer indeed extracts potential periodic dependencies, we visualize the feature similarity between each frame in a single gait sequence. As shown in  Figure~\ref{fig:heatmap}, the similarity between the features generated by the baseline method GaitPart and GaitGL does not have an obvious periodic trend, while the feature similarity of GaitFormer exhibit a periodic distribution. This demonstrates that the TPA module can indeed inject periodic prior to facilitate training. Furthermore, in order to reveal the learned periodicity of GaitFormer more intuitively, we display the original gait silhouettes along with the similarity in Figure~\ref{fig:oumvlp_vis}. Such results further demonstrate the effectiveness of GaitFormer.

\paragraph{Impact of the layer number in TAM}
TAM is incorporated to better decompose the temporal signals. Table~\ref{tab:tpm_ablation} shows results of TAM with different numbers of layers. The experiments are conducted on the OU-MVLP dataset. It can be observed that TAM with 6 layers achieves better performance compared with other variants while deeper TAM will bring high computational costs.

\paragraph{Impact of $T_d$ in AFPE} 
The component number of the AFPE is empirically set to 1 through our experiments. Here we provide the results on the OU-MVLP dataset in Figure~\ref{fig:ablation_FPE} with different component numbers  $T_d$ in AFPE. GaitFormer achieves the best performance when  $T_d$ is set to 1. This is consistent with the results we obtained by visualizing the OU-MVLP dataset,  where the period of gait silhouettes is about 30 frames. When $T_d$ is too large, the transform periodic transformer tends to overfit the complex AFPE.

\section{Conclusion}
In this work, we propose a plug-and-play strategy named Temporal Periodic Alignment (TPA) for gait recognition to make capital of the periodic inductive bias. Specifically, TPA encapsulates two key components, Adaptive Fourier-transform Position Encoding (AFPE) which injects the period prior, and Temporal Aggregation Module (TAM) which explores fine-grained temporal dependencies. We further build a simple yet strong baseline GaitFormer on the basis of TPA. Extensive experiments are conducted on several datasets to verify the effectiveness of these components.

\appendix
\section{More Details of Implementation}
\subsection{Framework of TAM}
Temporal Aggregation Module (TAM) is composed of Multi-Head Cross-Attention (MHCA) sub-layer and a position-wise feed-forward network (FFN), where the period-aware embeddings and tokens are directly sent into MHCA sub-layer to extract temporal information. The output feature computes temporal triplet loss as a temporal representation. At the same time, it is also sent into a fusion module to generate global features as classification representations. For the understanding of decomposition, inspired by time-series regression, input series are separated into trend-cyclical and seasonal parts, which reflect long-term progression and seasonality respectively. For gait recognition, the seasonal parts reflect the periodic template of the walking pattern, while the trend-cyclical parts mostly display shape and appearance.

\begin{figure}[htb] 
\centering 
\includegraphics[width=1\linewidth]{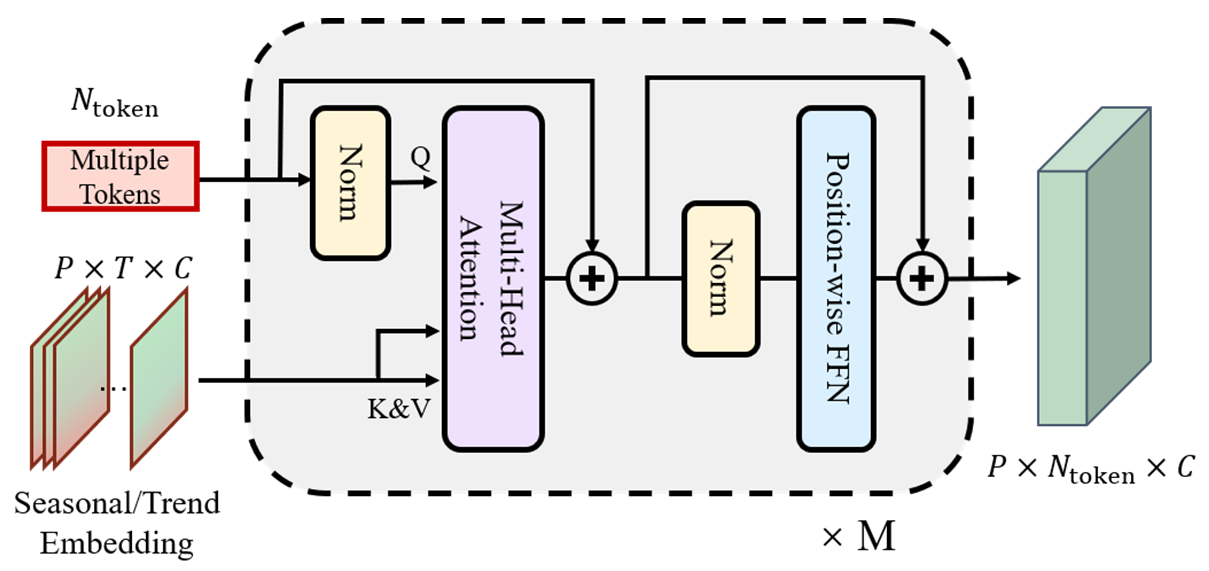 }
\caption{Framework of the TAM module.} 
\label{fig:TPM} 
\end{figure}
\subsection{Details of training objectives}
As mentioned, we replace the vanilla CE loss with the ArcFace loss \cite{deng2019arcface} to produce a larger margin. Marginal losses like ArcFace are widely adopted in face recognition and re-identification. The ArcFace loss is an improved version of cross-entropy loss, which introduces angular distance to constrain the distance between features of different categories. Thanks to the scale and angular margin, ArcFace loss achieves better generalization performance in face recognition and person re-identification tasks. Empirically We find that ArcFace brings faster and better convergence of experiments. The ArcFace loss can be formatted as:
\begin{equation}
\small
\label{arcface}
L_{cls} = -\frac{1}{N} \sum_{i=1}^{N} \log{\frac{e^{s \cdot(cos(\theta_{y_{i}} + m))}}{e^{s \cdot (cos(\theta_{y_{i}} + m))}+ \sum_{j=1,j\ne y_i}^{n} e^{s \cdot cos(\theta_j)}} }
\end{equation}
where $N$ and $n$ indicate batch size and class number, $s$ and $m$ are the scale and margin mentioned above, $y_i$ means the class index of the $i th$ instance, $\theta_j$ is the angle between the project layer weight and the $i_th$ feature. We set $s=32$ and $m=0.3$ in our experiments.

\begin{table}[t!]
  \small
   \centering
   \renewcommand{\arraystretch}{1.2}
  \tabcolsep=0.2cm
   \begin{tabular}{l|c}
 \toprule

          Ablation  & OU-MVLP \\
          \midrule
          \textbf{GaitFormer} & \textbf{91.60} \\ \midrule
          GaitFormer with Cross Entropy  & 91.20  \\
          GaitFormer with vanilla PE & 90.51\\ 
  \bottomrule
   \end{tabular}
  \caption{\small Ablation study of GaitFormer on OU-MVLP dataset. The number of transformer blocks is set as 6 in experiments. }
  \label{tab:pe_ablation}%
\end{table}

\begin{table*}[t!]
\begin{center}
\footnotesize

\renewcommand\arraystretch{1.0}
 \begin{tabular}{cccccccccccccc}
    \hline
     \multicolumn{1}{c}{\multirow{2}{*}{\textbf{Probe}}} &
     \multicolumn{1}{|c|}{\multirow{2}{*}{\textbf{Method}}} &
     \multicolumn{11}{c}{\textbf{Probe View}} & \multicolumn{1}{|c}{\multirow{2}{*}{\textbf{Mean}}} \\
    \cline{3-13}
  \multicolumn{1}{c}{} & \multicolumn{1}{|c|}{} & $0^{\circ}$     & $18^{\circ}$    & $36^{\circ}$    & $54^{\circ}$    & $72^{\circ}$    & $90^{\circ}$    & $108^{\circ}$   & $126^{\circ}$   & $144^{\circ}$   & $162^{\circ}$   & $180^{\circ}$ & \multicolumn{1}{|c}{} \\
    \midrule
    \multicolumn{1}{c|}{\multirow{8}{*}{NM}}     & \multicolumn{1}{l|}{GaitSet} & 90.8 & 97.9 & 99.4 & 96.9  & 93.6  & 91.7 & 95.0 & 97.8 & 98.9 & 96.8 & 85.8 & \multicolumn{1}{|c}{95.0}  
\\
\multicolumn{1}{l|}{}      & \multicolumn{1}{l|}{GaitPart} & 94.1  & 98.6  & 99.3  & 98.5  & 94.0    & 92.3  & 95.9  & 98.4  & 99.2  & 97.8  &  90.4 & \multicolumn{1}{|c}{96.2}  
\\
    \multicolumn{1}{l|}{}      & \multicolumn{1}{l|}{ReverseMask} & 96.5  & 98.4  & 99.2  & 98.0  & 97.1  & 95.5  & 97.4  & 99.2  & 99.3  & 99.1  & 95.0  & \multicolumn{1}{|c}{97.7}   
\\
    \multicolumn{1}{l|}{}      &  \multicolumn{1}{l|}{GaitGL}  & 96.0  & 98.3  & 99.0  & 97.9  & 96.9  & 95.4  & 97.0  & 98.9  & 99.3  & 98.8  & 94.0  & \multicolumn{1}{|c}{97.4}   
\\
  \multicolumn{1}{l|}{}  & \multicolumn{1}{l|}{3DLocal}  & {{96.0}}  & {{99.0}}  & 99.5  & 98.9 & 97.1 & 94.2 & 96.3 & 99.0 & 98.8 & 98.5 & 95.2  & \multicolumn{1}{|c}{97.5}  
\\
\multicolumn{1}{l|}{}  & \multicolumn{1}{l|}{GaitSet+}  &94.2 &  {98.2} & 99.7 & 97.5 & 95.6 & 94.9 & 95.8 & 97.7 & 98.1 & 97.5 & 93.3  & \multicolumn{1}{|c}{{{96.9}}}  
\\
  \multicolumn{1}{l|}{}  & \multicolumn{1}{l|}{GaitGL+}  & 96.3 & 98.5 & 99.2 &  98.3 & 98.4 & 97.7 &  \textbf{98.8} &  \textbf{99.8} &  \textbf{99.7} & 99.4 & 97.1  &\multicolumn{1}{|c}{{{98.5}}}  
\\

  \multicolumn{1}{l|}{}  & \multicolumn{1}{l|}{GaitFormer+}  & \textbf{97.0} & \textbf{100.0} & \textbf{100.0} &  \textbf{99.7} &   \textbf{98.8} &  \textbf{97.8} & 98.0 &  99.7 &   \textbf{99.7} &   \textbf{99.7} &
  \textbf{97.7} & \multicolumn{1}{|c}{{{ \textbf{98.9}}}}  
\\
    \midrule
  
    \multicolumn{1}{c|}{\multirow{8}{*}{BG}}     & \multicolumn{1}{l|}{GaitSet} & 83.8 & 91.2 & 91.8 & 88.8 & 83.3 & 81.0 & 84.1 & 90.0 & 92.2 & 94.4 & 79.0 &\multicolumn{1}{|c}{87.2}  
\\
\multicolumn{1}{l|}{}     & \multicolumn{1}{l|}{GaitPart} & 89.1 
    &94.8& 96.7& 95.1 & 88.3 & 84.9 & 89.0 & 93.5 & 96.1 & 93.8 & 85.8 & \multicolumn{1}{|c}{91.5}  
\\
    \multicolumn{1}{l|}{}      & \multicolumn{1}{l|}{ReverseMask} & 93.7 & 97.0 &  \textbf{97.3} & 95.8 &  \textbf{94.9}  &  \textbf{91.4} & 93.5 & 97.3 & 98.3 & 97.3 & 92.4 & \multicolumn{1}{|c}{ \textbf{95.3}}   
\\
    \multicolumn{1}{l|}{}      &  \multicolumn{1}{l|}{GaitGL} & 92.6 & 96.6 & 96.8 & 95.5 & 93.5 & 89.3 & 92.2 & 96.5 & 98.2 & 96.9 & 91.5  & \multicolumn{1}{|c}{94.5}   
\\
  \multicolumn{1}{l|}{}  & \multicolumn{1}{l|}{3DLocal}  & 92.9 & 95.9 & 97.8 &  \textbf{96.2} & 93.0 & 87.8 & 92.7 & 96.3 & 97.9 & 98.0 & 88.5  & \multicolumn{1}{|c}{94.3}  
\\
\multicolumn{1}{l|}{}  & \multicolumn{1}{l|}{GaitSet+}  &89.6 & 94.9 & 95.2 & 93.3 & 89.9 & 86.7 & 90.8 & 94.5 & 97.2 & 97.7 & 87.3& \multicolumn{1}{|c}{{{92.6}}}  
\\
  \multicolumn{1}{l|}{}  & \multicolumn{1}{l|}{GaitGL+}  & \textbf{94.0} & 96.7 & 97.0 &  \textbf{96.2} & 94.5 & 91.0 &  \textbf{93.7} &  \textbf{96.8} &  \textbf{98.6} &  \textbf{97.7} &  \textbf{92.6}  & \multicolumn{1}{|c}{{{ \textbf{95.3}}}}  
\\

  \multicolumn{1}{l|}{}  & \multicolumn{1}{l|}{GaitFormer+}  & 92.2 &   \textbf{97.7} & 97.1 & 95.8 & 94.1 & 87.6 & 93.1 & 96.5 & 98.1 & 97.5 & 92.3  & \multicolumn{1}{|c}{{{94.7}}}  
\\
    \midrule
    \multicolumn{1}{c|}{\multirow{8}{*}{CL}}     & \multicolumn{1}{l|}{GaitSet} & 61.4 & 75.4 & 80.7 & 77.3 & 72.1  & 70.1 & 71.5 & 73.5 & 73.5 & 68.4 & 50.0 & \multicolumn{1}{|c}{70.4}  
\\
  
\multicolumn{1}{l|}{}     & \multicolumn{1}{l|}{GaitPart} & 70.7 & 85.5 & 86.9 & 83.3 & 77.1 & 72.5 & 76.9 & 82.2 & 83.8 & 80.2 & 66.5 & \multicolumn{1}{|c}{78.7}  
\\
    \multicolumn{1}{l|}{}      & \multicolumn{1}{l|}{ReverseMask} &  \textbf{78.9} & 91.5 &  \textbf{93.1} & 91.1 & 85.6 & 81.0 &  \textbf{85.2} & 89.0 & 90.9 & 87.3 & 72.9  & \multicolumn{1}{|c}{86.0}   
\\
    \multicolumn{1}{l|}{}      &  \multicolumn{1}{l|}{GaitGL}  & 76.6 &  90.0 & 90.3 & 87.1 & 84.5 & 79.0 & 84.1 & 87.0 & 87.3 & 84.4 & 69.5  & \multicolumn{1}{|c}{83.6}   
\\
  \multicolumn{1}{l|}{}  & \multicolumn{1}{l|}{3DLocal}  & 78.2 & 90.2 & 92.0 & 87.1 & 83.0 & 76.8 & 83.1 & 86.6 & 86.8 & 84.1 & 70.9  & \multicolumn{1}{|c}{83.7}  
\\
\multicolumn{1}{l|}{}  & \multicolumn{1}{l|}{GaitSet+}  &72.0 & 86.3 & 89.7 & 87.6 & 80.2  & 79.1 & 80.9 & 82.2 & 86.5 & 78.7 & 69.6  & \multicolumn{1}{|c}{{{81.7}}}  
\\
  \multicolumn{1}{l|}{}  & \multicolumn{1}{l|}{GaitGL+}  &77.0 &  \textbf{93.2} & 92.9 & 90.5 & 83.8 & 81.3 & 85.0 & 89.7 & 90.1 & 86.0 & 73.9  & \multicolumn{1}{|c}{{{85.8}}}  
\\

  \multicolumn{1}{l|}{}  & \multicolumn{1}{l|}{GaitFormer+}  & 77.3 & 88.8 & 92.9 &  \textbf{93.1} &  \textbf{88.6} &  \textbf{82.1} &  \textbf{85.2} &  \textbf{91.0} &  \textbf{92.7} &  \textbf{90.5} &  \textbf{78.2}  & \multicolumn{1}{|c}{{ \textbf{87.3}}}  
\\
    \bottomrule
    
    \end{tabular}%
    \caption{Rank-1 accuracy (\%) of the compared method on CASIA-B under different probe views,excluding identical-view cases with the setting of large-scale training (LT, i.e., 74 subjects for training),.}\label{tab:casiab_results2}

\end{center}
\end{table*}

\begin{table}[htb]
  \small
   \centering
   \renewcommand{\arraystretch}{1.1}
  \tabcolsep=0.25cm
   \begin{tabular}{l|c|ccc}
 \toprule
          Methods & TPA & FLOPs & Params & Accuracy\\
          \midrule

          \multirow{2}{*}[-0.3ex]{GaitPart \cite{DBLP:conf/cvpr/FanPC0HCHLH20}} & \xmark  & 7.93G & 4.8M & 88.7 \\
           &$\checkmark$ & 8.41G & 9.8M & 89.4 \\
           \midrule
          \multirow{2}{*}[-0.3ex]{GaitGL \cite{DBLP:conf/iccv/LinZ021}} & \xmark  & 58.64G & 95.6M & 89.7 \\
           & $\checkmark$  & 59.44G & 100.6M & 90.4 \\
          \midrule

          \multirow{2}{*}[-0.3ex]{MetaGait* \cite{dou2022metagait}} & \xmark  & 58.64G & 95.8M & 89.1 \\
           & $\checkmark$  & 59.19G & 99.1M & 89.6 \\
  \bottomrule
   \end{tabular}
  \caption{\small The FLOPs and parameteres before and after we replace the Temporal Aggregator with our TPA strategy. It can be observed that our TPA strategy brings perceptible improvement, with small increase in calculation and parameters amount. MetaGait* represents our implementation of MetaGait, which is lower than the result in the origin paper.}
  \label{tab:plug2}%
\end{table}

\section{Additional Experiment Results}
In this section, we will give some additional detailed experimental results and related analyses. The results of other baselines are copied from their original papers.

\subsection{Visualization of AFPE feature}
\begin{figure*}[t!]
\centering

    \begin{subfigure}[b]{0.32\linewidth}
        \centering
        \includegraphics[width=\columnwidth]{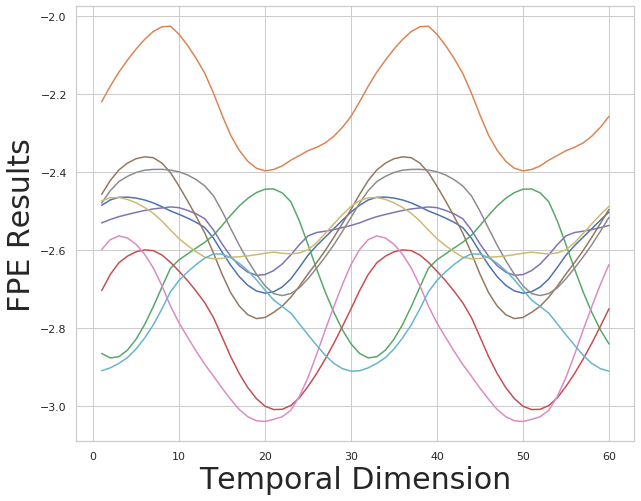} \\
        \includegraphics[width=\columnwidth]{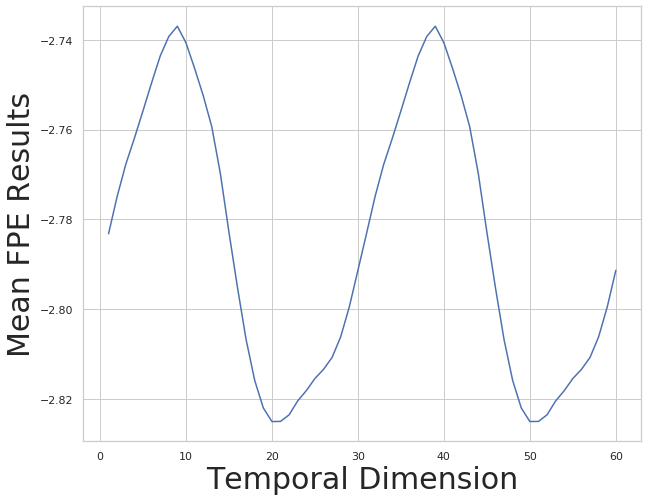}
        \caption{10 base sequences.}
    \end{subfigure}
   \begin{subfigure}[b]{0.32\linewidth}
        \centering
        \includegraphics[width=\columnwidth]{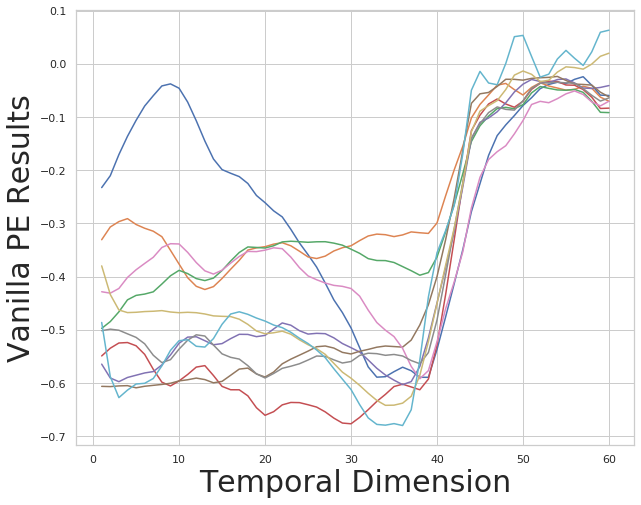} \\
        \includegraphics[width=\columnwidth]{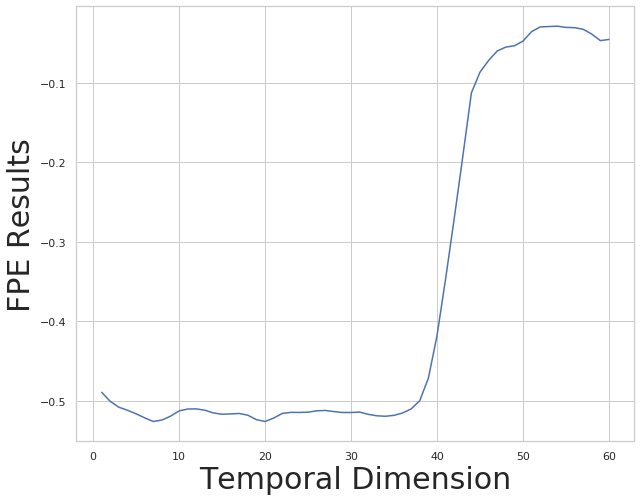}
        \caption{20 base sequences.}
    \end{subfigure}
    \begin{subfigure}[b]{0.32\linewidth}
        \centering
        \includegraphics[width=\columnwidth]{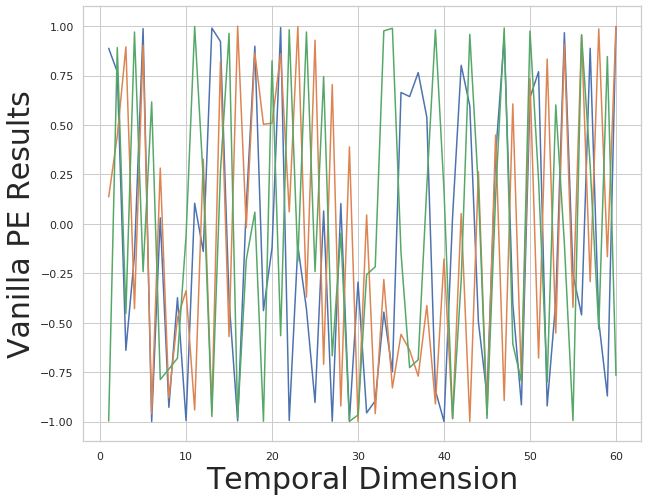} \\
        \includegraphics[width=\columnwidth]
        {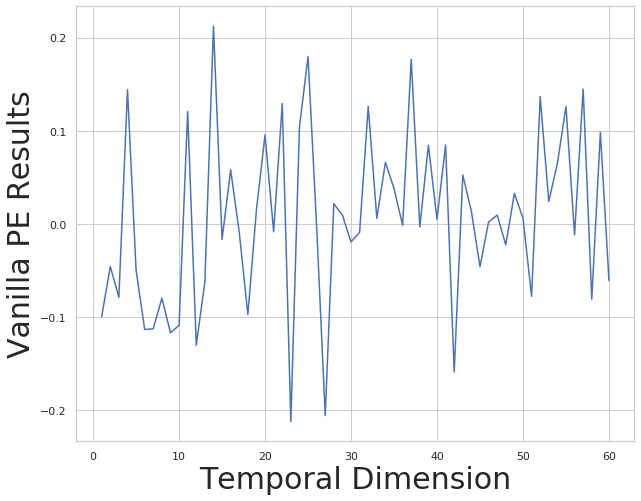}
        \caption{vanilla pe.}
    \end{subfigure}

\caption{AFPE and average amplitude of different experimental configurations of base sequences. If the number of base sequences is greater than 10, then 10 samples are randomly sampled from all sequences. }
\label{fig:FPE_viz}
\end{figure*}

The AFPE is introduced to encode a series of discrete time and extend the original signals into the positional-aware embeddings in an adaptive manner. Hence it is important to observe the final results generated by AFPE. The shape of the final AFPE is $1\times T \times C$, where $T$ is the total frames of the sequence on the temporal dimension and $C$ is the same as the input feature dimensions. The first row in Figure~\ref{fig:FPE_viz} shows 10 randomly selected base sequences and the second row shows the trend of the average AFPE result, which is the final AFPE added with embeddings. Such visualization is consistent with the assumption that the gait sequence is a composite of multiple similar 
periods.

\subsection{Additional results for GaitFormer}

\paragraph{Additional results under different Views on CASIA-B dataset}
We provide detailed results under different probe views on the CASIA-B dataset in Table~\ref{tab:casiab_results2}. There are 11 views ($0^\circ$,$18^\circ$, ......,$180^\circ$) under NM, BG and CL variants. It is worth noting that the number of samples of CASIA-B is too limited and easily overfitting with the transformer framework, so we utilize the model pre-trained on OUVMLP to finetune on CASIA-B dataset. Here methods with '+' denote two-stage training, warm-up on OUVMLP and continue the second stage of training and testing on CASIA-B dataset.

\paragraph{Additional ablation of GaitFormer Components}
we conduct more ablation experiments on the OU-MVLP dataset to explore the contribution of the modules. The experimental results are shown in Table~\ref{tab:pe_ablation}. 

We can observe that GaitFormer with vanilla CE bears slight performance degradation. Moreover, the vanilla position encoding (PE) regards each element as unique and our proposed AFPE force the network to exploit periodic prior. Table~\ref{tab:pe_ablation} compares the performance with AFPE and vanilla PE and the proposed AFPE is shown to excel, which demonstrates the effectiveness of our proposed AFPE.

\paragraph{Advantages of GaitFormer}
Compared with temporal normalization \cite{DBLP:journals/pr/LiaoYAH20} and phase synchronization \cite{5595941}. GaitFormer resorts to attention mechanisms to explicitly explore the temporal dependencies between different frames in the walking sequence. Such paradigms of temporal fusion are more intuitive and effective, which is displayed in our visualizations.

{\small
\bibliographystyle{ieee_fullname}
\bibliography{egpaper_230725}
}

\end{document}